\documentclass[final]{cvpr}
\usepackage{times}
\usepackage{epsfig}
\usepackage{graphicx}
\usepackage{amsmath}
\usepackage{amssymb}
\usepackage[resetlabels,labeled]{multibib}

\pdfoutput=1

\newcites{N}{References}

\usepackage{xcolor}

\graphicspath{{figures/}}

\newcommand{\Eq}[1]{Eq.~(\ref{eq:#1})}
\newcommand{\eq}[1]{\Eq{#1}}

\newcommand{\fig}[1]{Fig.~\ref{fig:#1}}
\newcommand{\tab}[1]{Table~\ref{tab:#1}}

\usepackage{graphicx}
\usepackage{amsmath}
\usepackage{amssymb}
\usepackage{booktabs}
\usepackage{comment}

\usepackage{mathtools}

\DeclarePairedDelimiter\abs{\lvert}{\rvert}%

\usepackage[pagebackref=true,breaklinks=true,colorlinks,bookmarks=false]{hyperref}

\def\cvprPaperID{****} % *** Enter the CVPR Paper ID here

\def\cvprPaperID{10136} % *** Enter the CVPR Paper ID here
\def\confYear{CVPR 2021}

\begin{document}
%%%%%%%% TITLE - PLEASE UPDATE
\title{No frame left behind: Full Video Action Recognition}
%\title{Do we need more frames for action recognition?}  % **** Enter the paper title here
% Other title options here: ...........

\author{Xin Liu$^1$
\ \ Silvia L. Pintea$^1$ \ \ Fatemeh Karimi Nejadasl$^2$ \ \ Olaf Booij$^2$ \ \ Jan C. van Gemert$^1$\\ 
Computer Vision Lab, Delft University of Technology$^1$
\ \ \ \ \ \ \ \ \ TomTom$^2$
}

\maketitle
%\thispagestyle{empty}
%%%%%%%%% BODY TEXT - ENTER YOUR RESPONSE BELOW
\begin{abstract}
Not all video frames are equally informative for recognizing an action. 
It is computationally infeasible to train deep networks on all video frames when actions develop over hundreds of frames.
A common heuristic is uniformly sampling a small number of video frames and using these to recognize the action. 
Instead, here we propose \emph{full video action recognition} and consider all video frames. 
To make this computational tractable, we first cluster all frame activations along the temporal dimension based on their similarity with respect to the classification task, and then temporally aggregate the frames in the clusters into a smaller number of representations.
Our method is end-to-end trainable and computationally efficient as it relies on temporally localized clustering in combination with fast Hamming distances in feature space. 
We evaluate on UCF101, HMDB51, Breakfast, and Something-Something V1 and V2, where we compare favorably to existing heuristic frame sampling methods.
\end{abstract}
\section{Introduction}

\begin{figure}[t]
    \centering
    \includegraphics[width=0.47\textwidth]{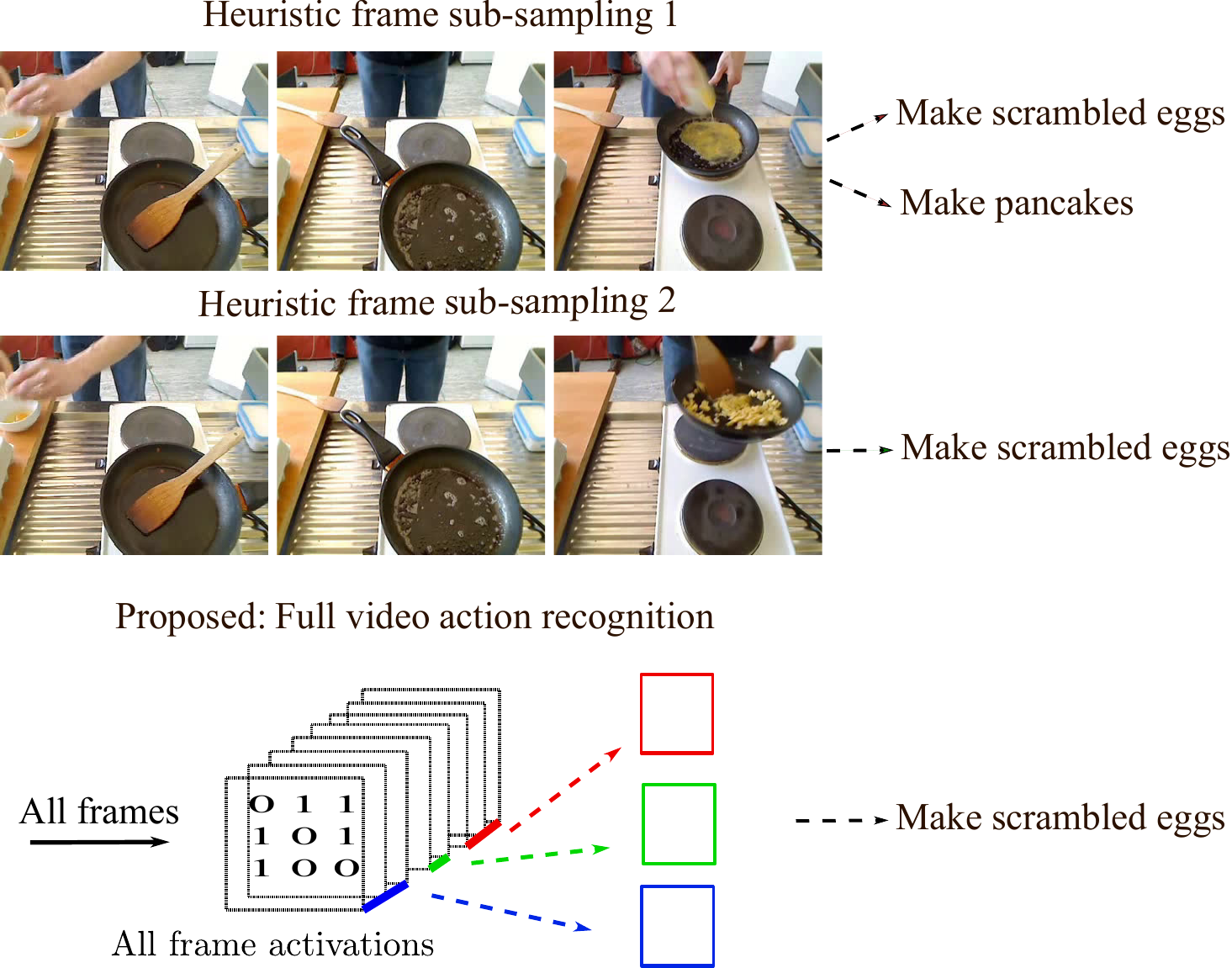}
    \caption{
    Sub-sampling can miss crucial frames in videos and may cause confusion for action recognition: e.g. compare the two sub-samplings heuristics in row 1 and row 2: Without sampling the dish in the pan it is difficult to classify. 
    Instead, as shown in row 3, we propose to efficiently use all frames during training by clustering frame activations along the temporal dimension and aggregating each cluster to a single representation. 
    The temporal clustering is based on Hamming distances over frame activations, which is computationally fast.
    With the assumption that similar activations have similar gradients, the aggregated representations approximate the individual frame activations. 
    We efficiently utilize all frames for training without missing important information. 
    }
    \label{fig:concept}
\end{figure}

% No sub-sampling; Use all frames
Videos have arbitrary length with actions occurring at arbitrary moments.
Current video recognition methods use CNNs on coarsely sub-sampled frames~\cite{carreira2017quo,korbar2019scsampler,lin2019tsm,ren2020best,sudhakaran2020gate,wang2016temporal, wu2019multi,Wu_2019_CVPR,yeung2016end, zolfaghari2018eco} because using all frames is computationally infeasible.
Sub-sampling, however, can miss crucial frames for action recognition.
For example, as shown in~\fig{concept}, sampling the frame with the dish in the pan is crucial for correct recognition. 
We propose to do away with sub-sampling heuristics and argue for leveraging all video frames: Full video action recognition.

% What is the computatational problem? Memory.
It is worth analyzing why training CNNs on full videos is computationally infeasible in terms of memory and calculations. The calculations in the forward pass yield activations, while the backward pass calculations give gradients which are summed over all frames to update the weights. Many of these calculations can be done in parallel and thus are well-suited for modern GPUs. When treating videos as a large collection of image frames, the amount of calculations are not too different from those on large image datasets~\cite{ImageNet}. Regarding memory, however, there is a crucial difference between videos and images: The video loss function is not per-frame but on the full video. Hence, to do the backward pass, all activations for each frame, for each filter in each layer need to be stored in memory. This even doubles for storing their gradients. With 10-30 frames per second, this quickly becomes infeasible for even just a few minutes of video. Existing approaches can trade off memory for compute~\cite{chang2018reversible, chen2016training, gomez2017reversible} by not storing all intermediate layers, yet they do not scale to video as they would still need to store each frame. The main computational bottleneck for training video CNNs is memory for frame activations.

% Summing activations is OK when linear; thus cluster linear parts which can be summed
Here, we propose an efficient method to use all  video frames during training. The forward pass computes frame activations and the backward pass sums the gradients over the frames to update the weights.
Now, if only the network was linear, then a huge memory reduction could be gained by first summing all frame activations in the forward pass, which would reduce to just a single update in the backward pass. Yet, deep networks are infamously non-linear, and have non-linearities in the activation function and in the loss function. Thus, if all frames were independent, treating the non-linear network as linear would introduce considerable approximation errors. However, subsequent frames in a video are strongly correlated, and it's this correlation that makes it possible for existing approaches to use sub-sampling. Instead of sub-sampling, we propose to process all frames and exploit the frame correlations to create groups of frames where the network is approximately linear. We use the $ReLU$ (Rectified Linear Unit) activation function, which is linear when the signs of two activations agree, to estimate which parts of the video are approximately linear. This allows us to develop an efficient clustering algorithm based on Hamming distances of frame activations as illustrated in~\fig{concept}. By then aggregating the approximately linear parts in a video in the forward pass, we make large memory savings in the backward pass while still approximating the full video gradient.

% Contribution
% We summarize the contributions of our work as follows:
% (i) We propose a method that allows us to use most or even all video frames for training action recognition by approximated individual frame gradients with the gradients of temporally aggregated frame activations;
% (ii) We devise an end-to-end trainable approach for efficient grouping of video frames based on temporally localized clustering and Hamming distances;
% (iii) Extensive experiments demonstrate that our method compares well to state-of-the-art methods on several benchmark datasets such as UCF101, HMDB51, Breakfast, and Something-Something V1 and V2.

\noindent
We summarize the contributions of our work as follows:
\begin{itemize}
\item We propose a method that allows us to use most or even all video frames for training action recognition by approximated individual frame gradients with the gradients of temporally aggregated frame activations;
\item We devise an end-to-end trainable approach for efficient grouping of video frames based on temporally localized clustering and Hamming distances;
\item Extensive experiments demonstrate that our method compares well to state-of-the-art methods on several benchmark datasets such as UCF101, HMDB51, Breakfast, and Something-Something V1 and V2.
\end{itemize}

\section{Related work}
%----------------------------------------------------

%In this paper we use the terms activations and activations interchangeably. 
% JvG: I compacted/focused related work; original below in comments

\textbf{Action recognition architectures.} Actions in video involve motion, leading to deep networks which include optical flow~\cite{ feichtenhofer2016convolutional, gammulle2017two, simonyan2014two}, 3D convolutions~\cite{carreira2017quo, du2014c3d, hara2018can, ji20123d} and recurrent connections~\cite{gammulle2017two,singh2016multi,veeriah2015differential,ullah2017action, wu2019liteeval}. 
Instead of heavy-weight motion representations, a single 2D image can reveal much of an action~\cite{jain2015objects4action,KarpathyCVPR14, simonyan2014two,wang2016temporal}. 2D CNNs are extremely efficient, and by adding motion information by concatenating a 3D module in ECO~\cite{zolfaghari2018eco}, modeling temporal relations in TSN~\cite{zhou2018temporal} or simply shifting filter channels over time in TSM~\cite{lin2019tsm} their efficiency is complemented by good accuracy. For this reason, we build on the TSM~\cite{lin2019tsm} architecture and modify it for full video action recognition. 

\textbf{Frame sampling for action recognition.}
% sampling even segments
Realistic videos contain more frames than can fit in memory.
To address this, current methods train by using sub-sampled video frames~\cite{carreira2017quo,lin2019tsm,wang2016temporal,zolfaghari2018eco}.
Additionally, the SlowFast~\cite{feichtenhofer2019slowfast} network also explores the resolution trade-off across temporal, spatial and channel dimension. Rather than using a fixed frame sampling strategy, the sampling can be adaptive~\cite{lin2019tsm,sudhakaran2020gate,wu2019multi,Wu_2019_CVPR,yeung2016end}, or
learned to select the best frame \cite{ren2020best}, or rely on clip sampling~\cite{korbar2019scsampler}.
In our work we do not sub-sample frames, but use all frames of the videos, however our clustering is adaptive as it dynamically adapt to the task and the loss function.

Using a subset of frames is computationally more efficient.
Using 5-7 frames is sufficient for state-of-the-art action recognition on short videos \cite{schindler2008action}.
Aiming for training efficiency, the work in \cite{wang2019e2} uses stochastic mini-batch dropping which drops complete batches rather than frames, with a certain probability.
Similarly, \cite{wu2020multigrid} uses variable mini-batch shapes with different spatio-temporal resolutions varied according to a schedule.
Unlike these methods, we do not focus on training efficiency, but propose a method that allows the network to see all video frames during training. 

%----------------------------------------------------
\textbf{Temporal pooling.}
To integrate frame-level features, TSN~\cite{wang2016temporal} uses average pooling in the late layers of the network.
ActionVLAD~\cite{GirdharActionVLAD2017} integrates two-stream networks with a learnable spatio-temporal feature aggregation.
Instead of performing temporal pooling or aggregation at a late stage of the network,
in \cite{fernando2016rank} RankSVM is used to rank frames temporally and then pool them together. 
As a follow-up, in~\cite{BilenDynamicImage2018} a 'dynamic image' is introduced, which is a compact representation of the videos frames using the `rank pool' operation. 
In \cite{sener2020temporal,song2018deep} temporal aggregation via pooling and attention is used. 
Similar to these methods, our proposal performs a temporal pooling of the network activations, however this aggregation is done over clustered activations and it allows us to process all video frames.  

%----------------------------------------------------
\textbf{Efficient backpropagation.}
Given that 2/3 of the training computations and memory are spent in the backward pass, existing work focuses on approximations.
It is more memory efficient to recompute activations from the previous layer instead of storing them~\cite{gomez2017reversible}, however this comes at the cost of increased training time.
In \cite{malinowski2020sideways} gradient approximations are used where activations are overwritten when new frames are seen without waiting for the backward pass to be performed. 
Also for efficient backpropagation, randomized automatic differentiation can be used~\cite{oktay2020randomized}, gradients can be reused during training~\cite{goli2020resprop}, or even quantized during backpropagation~\cite{wiedemann2020dithered}.
Similar to these works, we use all frames to approximate the full video gradient.

\section{Aggregated temporal clusters}

%---------------------------------------------------------------------------
\subsection{Approximating gradients}
We enable the use of all frames of a video during training.
To this end, we calculate a single gradient to approximate the gradients of a group of frames.
Our hypothesis is that nearby frames in a video are alike, and thus have similar activations, leading to congruent updates.
When using the $ReLU$ (Rectified Linear Unit) activation function, we know that for activations with agreeing signs, the activation function is linear.
Assuming that similar frames are approximately linear, the standard computation of the sum of gradients over all frames, becomes equivalent to first summing all frame activations and then computing a single gradient.
This is computationally and memory efficient.
Mathematically, for frames $i$, this can be formulated as:
\begin{equation}
    \sum_i \nabla_{\mathbf{w}} L(h(\mathbf{x}_i \mathbf{w})) = 
        \nabla_{\mathbf{w}} L\left(\sum_i h( \mathbf{x}_i \mathbf{w}) \right),
        \label{eq:ideal}
\end{equation}
where $\mathbf{x}$ are frame activations, $\mathbf{w}$ are the network weights, $h(\cdot)$ is an activation function, and $L(\cdot)$ is the loss function.
Note that~\eq{ideal} only holds in the ideal case when the activation function $h$ is linear for similar frames and the loss function $L$ is also linear.
This is not generally the case, and this approximation introduces an error.

With the above ideal scenario in mind, we can use all video frames without calculating the gradient for each frame, by grouping frames that agree in the sign of their activations $\mathbf{x}$.
Over these grouped activations we calculate a single gradient $\nabla_{\mathbf{w}}L( \sum_{i} h (\mathbf{x}_i \mathbf{w}))$.
However, for similar frames the sign of their activation values may not be in complete agreement.
Therefore, we aim to find which frames can be safely grouped together, to minimize the error introduced by our approximation in~\eq{ideal}.

%---------------------------------------------------------------------------
\subsection{Error bound for the approximation}
\begin{figure*}
    \centering
    \includegraphics[width=0.9\textwidth]{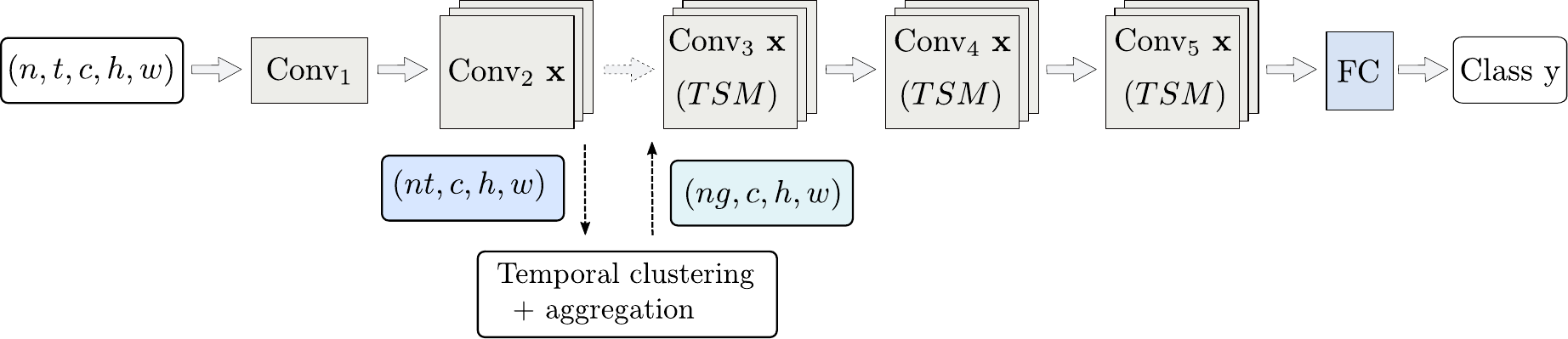}
    \caption{We adopt 2D ResNet-50 with TSM~\cite{lin2019tsm} a backbone. 
    The input batch size is $n$ with $t$ frames. 
    We cluster the activations of the first block of size $(n t, c, h, w)$ which groups $t$ frames into $g$ clusters and outputs new activations of size $(n g, c, h, w)$, as input to the next network blocks. 
    Our full video method efficiently utilizes all frames and is end-to-end trainable.
    }
    \label{fig:framework}
\end{figure*}
\begin{figure}
    \centering
    \includegraphics[width=0.46\textwidth]{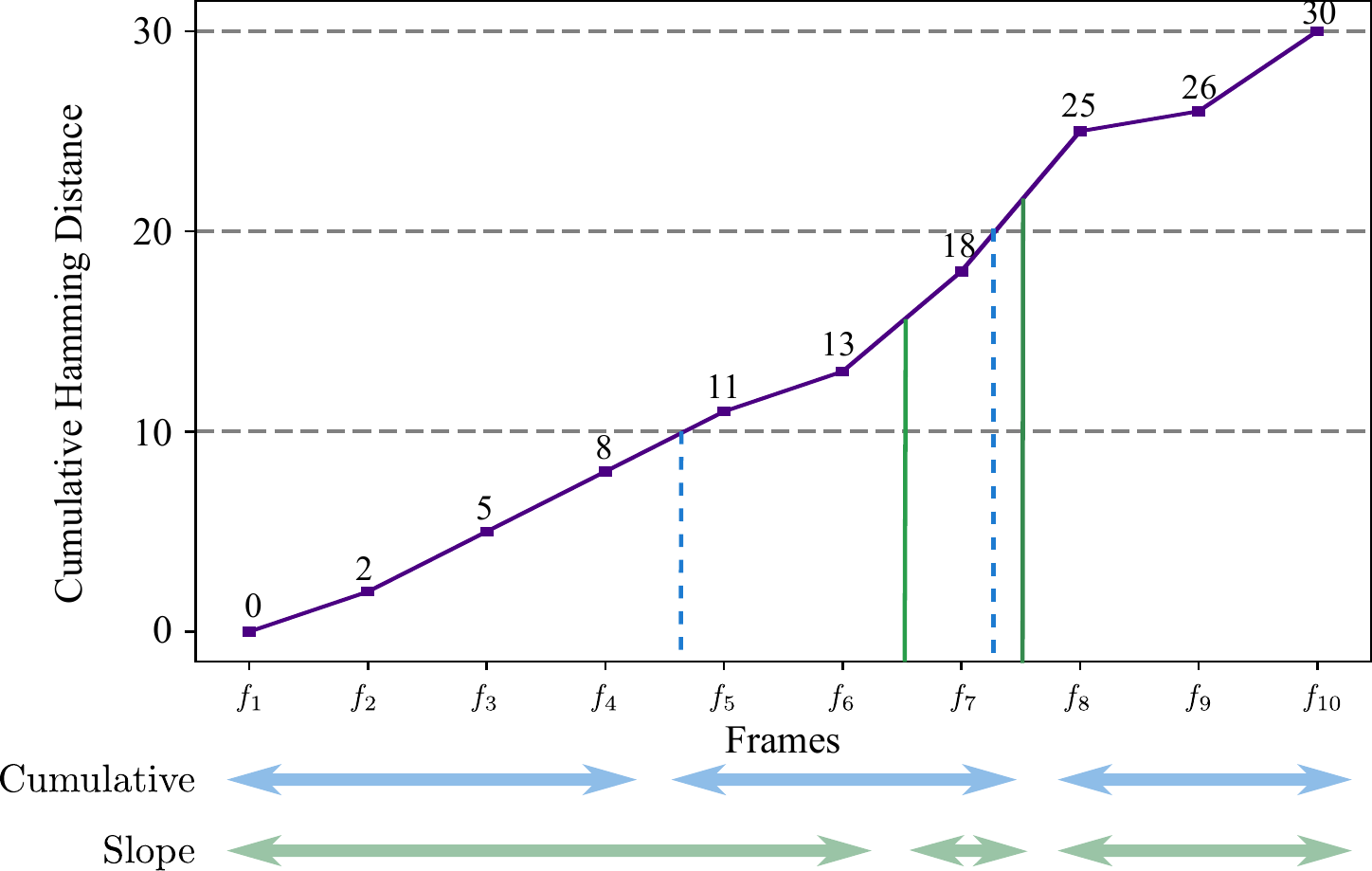}
    \caption{An illustration of our two clustering algorithms. 
    The numbers on the solid line are pair-wise Hamming distances and the solid line is the cumulative Hamming distance of frame $f_1$ to $f_{10}$. 
    For $g{=}3$ clusters, the \emph{cumulative clustering} groups frames by dividing the total cumulative distance on the $y$-axis into 3 equally distanced segments, as shown with the dashed lines resulting in the 3 clusters $(f_1{-}f_4)$, $(f_5{-}f_7)$ and $(f_8{-}f_{10})$.
    The \emph{slope clustering} algorithm is based on the slope of the curve and here selects the top-2 largest slopes, as shown with the solid green lines, which results in the 3 clusters: $(f_1{-}f_6)$, $(f_7)$, $(f_8{-}f_{10})$. }
    \label{fig:cumulative clustering}
\end{figure}
\begin{figure}
    \centering
    \includegraphics[width=0.47\textwidth]{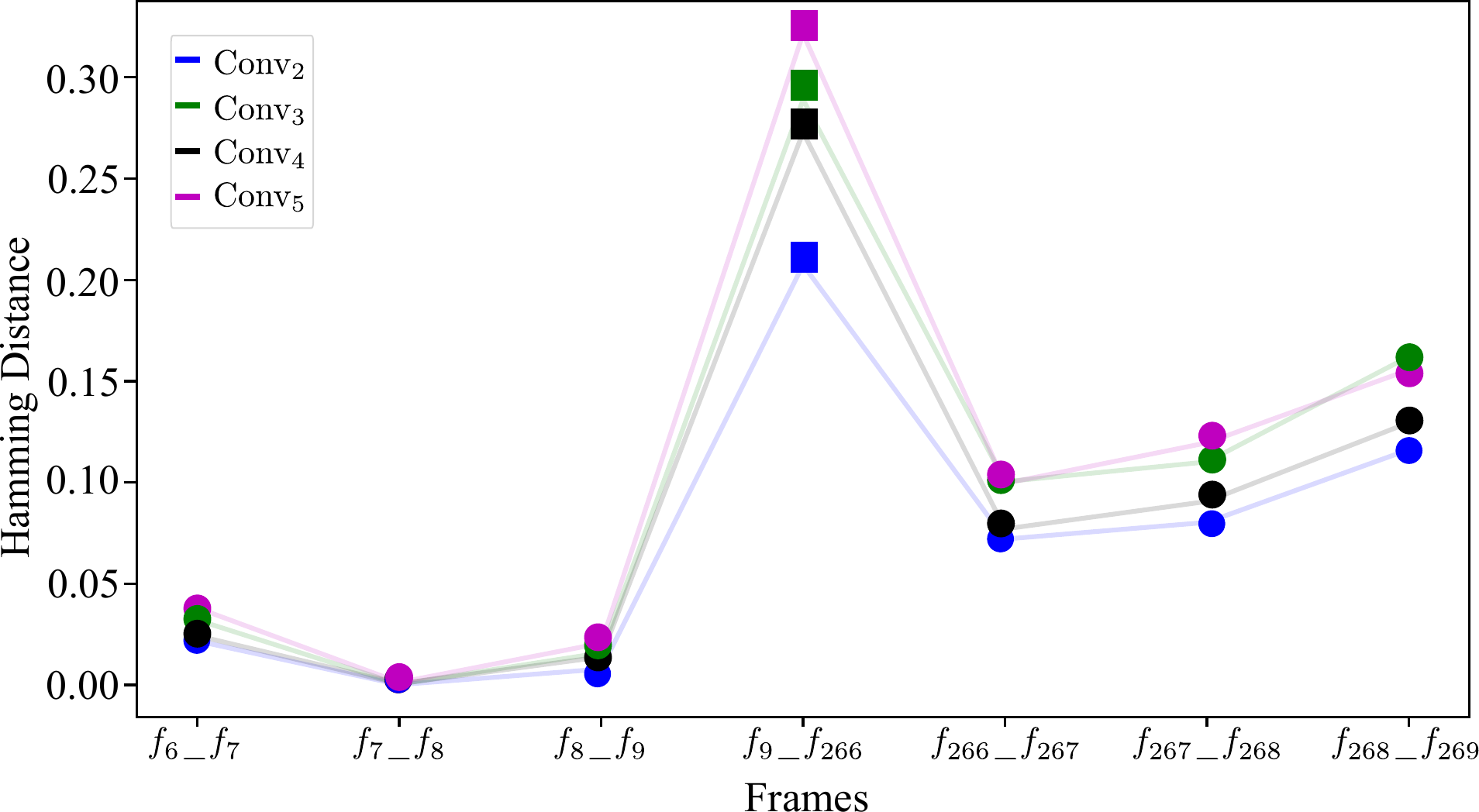}
    \caption{Hamming distances between similar frames and dissimilar frames across 4 blocks of ResNet. 
    The frames are taken from a single Breakfast~\cite{Breakfast} video. 
    We denote the frames that are similar to their neighbors with circles and the dissimilar ones with squares.
    Hamming distances are consistent across blocks.
    }
    \label{fig:dist_layers}
\end{figure}

For ease of explanation, we consider two input video frames and their activations $\mathbf{x}{=}\{\mathbf{x}_{1}, \mathbf{x}_{2} \}$, and a convolutional operation with parameters $\mathbf{w}$, denoted by $\mathbf{x} \mathbf{w}$.
The two frames have the same class label, $y$, since they come from the same video.
We consider a multi-class setting using the cross-entropy loss in combination with the softmax function $q$, which for these two samples is:
%\jvg{For eq 2, this $\frac{1}{2}$ in front of the loss is not there in eq 1. Option 1: average it also in eq 1 (but this is not what really happens I guess?); option 2: remove the $\frac{1}{2}$ in eq 2, 3, 4, 5, 6. Which may clean it up a bit, as $\frac{1}{2}$ doesn't really have a function? \slp{In eq 1 it does not need to be outside, itś inside the $\nabla_{\mathbf{w}} L$. If we remove it then eq 6 does not apply. We need the normalization /2.}}:
\begin{align}
    L(\mathbf{x}, y) &= -\frac{1}{2} \left( \log q_y(\mathbf{x}_{1}) + \log q_y(\mathbf{x}_{2}) \right),
    \label{eq:loss}
\end{align}
where 
$q_c(\mathbf{x}_{i}) {=} \frac{\exp(h(\mathbf{x_{i}}\mathbf{w}_c))}{\sum_{j=1}^C\exp( h(\mathbf{x_{i}}\mathbf{w}_j) )}$,
$c{\in}\{1, .., C\}$ indexes video classes and $h(\cdot)$ is the $ReLU$ activation function. 
The gradient of the loss with respect to $\mathbf{w}$ is:
\begin{align}
\nabla_{\mathbf{w}} L(\mathbf{x}, y) = 
        \frac{\mathbf{x}_1  ( q_c (\mathbf{x}_{1})  - \delta_{yc} ) + \mathbf{x}_2  ( q_c (\mathbf{x}_{2})  - \delta_{yc} )}{2},  \label{eq:grad_orig}
\end{align}
where $\delta_{yc}$ is the Dirac function which is 1 when $c{=}y$.
In our method, we first average the two activations after the convolution and before the $ReLU$.
We can do this, because if we assume the activations have agreeing signs $\text{sign}(\mathbf{x}_1\mathbf{w}){=}\text{sign}(\mathbf{x}_2\mathbf{w})$, then it holds that: $\frac{ h(\mathbf{x}_1\mathbf{w}) + h(\mathbf{x}_2\mathbf{w})}{2} = h(\frac{ \mathbf{x}_1\mathbf{w} + \mathbf{x}_2\mathbf{w}}{2})$.
In this case the cross-entropy loss becomes: 
\begin{equation}
    \hat{L}(\mathbf{x}, y){=}-\log q_y \left(\frac{\mathbf{x}_1 + \mathbf{x}_2}{2} \right). 
\end{equation}
In the backward pass, we calculate a single gradient of the averaged activations as follows:
\begin{align}
\nabla_{\mathbf{w}} \hat{L}(\mathbf{x}, y) &= 
        \frac{\mathbf{x}_1+\mathbf{x}_2}{2}   
        \left( q_c\left(\frac{\mathbf{x}_1+\mathbf{x}_2}{2}\right) - \delta_{yc}  \right),      \label{eq:grad_our}
\end{align}

We now estimate the relative error introduced by our approximation by comparing equations~\eq{grad_orig} and~\eq{grad_our} using Jensen's inequality. 
%, which in general case for a convex function $f$ is:  
%$f \left( \frac{ \sum_i a_i \mathbf{x_i} }{\sum_i a_i} \right)  \le \frac{ \sum_i a_i f(\mathbf{x}_i) }{\sum_i a}$.
We start from the softmax function $q_c(\cdot)$ and we recover back equations \eq{grad_orig} and \eq{grad_our}. % where the sample weights $a_i$ are $1$
The softmax function $q_c(\cdot)$ is convex, therefore we can apply to it Jensen's inequality for the samples $\mathbf{x}_1$ and $\mathbf{x}_2$: $q_c\left(\frac{(\mathbf{x}_1 + \mathbf{x}_2)}{2}\right)  \le \frac{q_c(\mathbf{x}_1) + q_c(\mathbf{x}_2)}{2}$. 
We start by considering the case $\frac{(\mathbf{x}_1 + \mathbf{x}_2)}{2} > 0$. If we multiply both sides of this inequality with $\frac{(\mathbf{x}_1 + \mathbf{x}_2)}{2}$ we obtain that:
\begin{align}
    \frac{(\mathbf{x}_1 + \mathbf{x}_2)}{2} q_c & \left( \frac{ (\mathbf{x}_1 + \mathbf{x}_2)}{2}\right) \le \frac{ \mathbf{x}_1 q_c(\mathbf{x}_1) + \mathbf{x}_2 q_c(\mathbf{x}_2)}{2} \nonumber \\ 
    & - \frac{1}{4}(\mathbf{x}_1 - \mathbf{x}_2)(q_c(\mathbf{x}_1) - q_c(\mathbf{x}_2)).
    \label{eq:ineq}
\end{align}
In the left hand side of the inequality we recover precisely the $\nabla_{\mathbf{w}} \hat{L}(\mathbf{x}, y)$ given by \eq{grad_our}, while in the right hand side we recover~\eq{grad_orig} minus the approximation error as $\nabla_{\mathbf{w}} L(\mathbf{x}, y) - \frac{1}{4} (\mathbf{x}_1 - \mathbf{x}_2)(q_c(\mathbf{x}_1) - q_c(\mathbf{x}_2))$. Note that for the case $y{=}c$ the additional Dirac terms in $\mathbf{x}$ cancel out.
We now consider also the case $\frac{(\mathbf{x}_1 + \mathbf{x}_2)}{2} \le 0$, which together with the previous case leads to the following bound on the absolute difference between the gradients in \eq{grad_orig} and \eq{grad_our}:
\begin{equation}
\small
\abs{ \nabla_{\mathbf{w}} L(\mathbf{x}, y) - \nabla_{\mathbf{w}} \hat{L}(\mathbf{x}, y)} \le 
\frac{1}{4} \abs{ (\mathbf{x}_1 - \mathbf{x}_2)(q_c(\mathbf{x}_1) - q_c(\mathbf{x}_2))}.
\end{equation}
Thus, the difference between the two gradient updates is bounded by a function depending on the difference between the activations and their softmax responses. 
The closer to 0 the difference between the activations the smaller the difference between their gradient updates.
We show in the experimental section that, indeed, small differences in the activations entail small differences in the loss.
%\xl{we show gradients in fig 5?}
% It is bad practice to refer to experiments in the method. It confuses the reader.

The inequality in~\eq{ineq} holds under the condition that the sign of activations agree. 
Therefore, we want to group frames based on the sign similarity of their activations.

%---------------------------------------------------------------------------
\subsection{Temporal clustering and aggregation}
% Under the assumption that processing more video frames improves action recognition accuracy, \jvg{it is possible even if the assumption is wrong :) }
Using our proposal in~\eq{grad_our} allows training on all video frames.
We group frames based on the sign agreement of their activations. 
An efficient way to do this, is to binarize the activation values by using the $sign$ function and
compute a fast Hamming distance between binarized activations to determine which frames to group.

Consecutive frames in a video are more likely to be similar in appearance and are thus more likely to have similarly signed activations.
Therefore, we explore two variants of a temporal clustering algorithm based on Hamming distances, where we allow a fixed number of clusters $g$ to match the available memory. 
We employ the temporal order of video frames and calculate Hamming distances only with neighboring frames.
\fig{cumulative clustering} illustrates the two temporal clustering algorithms we consider here: \emph{cumulative clustering} and \emph{slope clustering}.
We start by calculating the cumulative Hamming distance $\mathcal{C}(\mathbf{x})$ for neighboring frames along the temporal order:
\begin{equation}
    \mathcal{C}_N(\mathbf{x}) = \sum_{\substack{i=1}}^{N-1} H(\mathbf{x}_i, \mathbf{x}_{i+1}),
\end{equation}
where $\mathbf{x}_i$ is the binarized activation of frame $i$, $H(\cdot, \cdot)$ is the Hamming distance, and $N$ is the total number of frames.
For \emph{cumulative clustering}, we divide the total cumulative Hamming distance, %\fkn{from here to the end of paragraph can be removed. The sentence is already clear and formula looks strange to me :-)} \jvg{I would prefer to make it precise with an equation in addition to the text :-) } 
$\mathcal{C}(\mathbf{x})$,  into $g$ even segments, where the cluster id for frame $i$ is $ \lceil g \frac{\mathcal{C}_i( \bf{x})}{ \mathcal{C}_N(\bf{x})}  \rceil$.
%where the $k$-th segment boundary is: $\{f \in \{1,..N\} \mid f = \argmin_{i \in \{1,..N\}} | C_i (\mathbf{x}) - k \frac{C_N (\mathbf{x})}{g}) | \}$. \slp{Is this correct?} \jvg{I would say: 
For the \emph{slope clustering}, the boundaries of the segments are defined by the frame indexes corresponding to the top-$g$ largest slopes  where the cumulative distance increases the most. % $\frac{dC_N}{ d\mathbf{x}}(\mathbf{x})$
%If we denote the set of he top-g largest slopes by $\text{top}_g(\mathbf{x})$, then the frame boundaries are the frames whose slopes are closest to one of these top-g slopes:  $\{f \in \{1,..N\} \mid \exists j\in \{1,..g\}, f = \argmin_{i \in \{1,..N\}} | \frac{dC_i}{ d\mathbf{x}}(\mathbf{x}) - \text{top}_{j}(\mathbf{x}) | \}$. \slp{Is this correct?}

For efficiency, we cluster early on in the network, and input to the subsequent layers only aggregated activations.
We assume that the sign of the activations corresponding to two similar frames, approximately agree throughout the network.
To validate this, we visualize in~\fig{dist_layers} the Hamming distance over activations corresponding to similar and dissimilar frames. 
The distances corresponding to similar frames remain consistent across different layers.

\clearpage

Putting everything together, we input a set of $n$ videos into our TSM-based~\cite{lin2019tsm} network architecture. 
After the first block, we apply temporal clustering and average the activations within each cluster, giving rise to $g$ activations per video.
These aggregated activations are input to the subsequent blocks of the network.
Our method efficiently utilizes all frames for training and it is end-to-end trainable, as the gradients propagate directly through the aggregated feature-maps.
\fig{framework} depicts the outline of our method.
\section{Experiments}
We evaluate our method on the action recognition datasets Something-Something V1 \& V2~\cite{goyal2017something}, UCF-101~\cite{soomro2012dataset}, HMDB51~\cite{Kuehne2011HMDBAL} and Breakfast~\cite{Breakfast}.
The consistent improvements show the effectiveness and generality of our method.
We validate and analyze our method on a fully controlled Move4MNIST dataset we created.
We also include ablation studies of the components of our method.
%All experiments are conducted with the single modality RGB frames and evaluated on the validation set.

%---------------------------------------------------------------------------
\smallskip\noindent \textbf{Datasets}. 
%\slp{If we need more space: the Move4MNIST can move to sec 4.1 and the rest can go away.} 
Something-Something V1~\cite{goyal2017something} consists of 86k training videos and 11k validation videos belonging to 174 action categories.
The second release V2 of Something-Something increase the number of videos to 220k.
The UCF101~\cite{soomro2012dataset} dataset contains 101 action classes and 13,320 video clips.
The HMDB51~\cite{Kuehne2011HMDBAL} dataset is a varied collection of movies and web videos with 6,766 video clips from 51 action categories.
Breakfast~\cite{Breakfast} has long videos of human cooking activities with 10 categories with 1,712 videos in total, with 1,357 for training and 335 for testing.
Our fully controlled dataset Move4MNIST has four action classes \emph{\{move up, move down, move left, move right\}}, and 1,800 videos for training and 600 videos for testing.
Each video has 32 frames, with a digit from MNIST~\cite{lecun2010mnist} moving on a UCF-101 video background. To obtain a per-frame ground truth of which frames are relevant we randomly inserted a consecutive chunk of UCF-101 background frames, black frames and frames with MNIST digits that are not part of the target classes. An example is shown in~\fig{clustering_compare}.

%---------------------------------------------------------------------------

\smallskip\noindent \textbf{Training \& Inference}. 
Following the setting in TSM~\cite{lin2019tsm}, our models are fine-tuned from Kinetics~\cite{kay2017kinetics} pre-trained weights and we freeze the Batch Normalization~\cite{ioffe2015batch} layers for HMDB51~\cite{Kuehne2011HMDBAL} and UCF101~\cite{soomro2012dataset} datasets.
For other datasets, our models are fine-tuned from ImageNet~\cite{ImageNet} pre-trained weights.
To optimize the GPU we train with a fixed number of frames per batch. If the video has less frames, we pad it repeatedly with the last video frame. 
We compare and cluster the activations of all the frames in each video, and get $g$ average activations for each video, from the first block of our model.
We set the number of clusters to $g=\{8,16\}$  to align with the sub-sampling methods using 8 or 16 frames.
%The $g$ average activations of each video are passed to the second block of our model for training.
During testing, we follow the setting of TSM and sample one clip per video and use the full resolution image with the shorter side 256.

%---------------------------------------------------------------------------
\smallskip\noindent \textbf{Backbone architecture.}
For a fair comparison with the state-of-the-art, we evaluate our method on the TSM~\cite{lin2019tsm} backbone replying on the ResNet-50~\cite{he2016deep} architecture.
We use TSM with a ResNet-18 as the backbone for the experiments on our toy dataset Move4MNIST and for model analysis on the Breakfast dataset.

%=====================================================================================
\subsection{Are more frames better?}

To make it computationally possible to use all individual frames we evaluate on the fully controlled Move4MNIST to test if using more frames during training is better than sub-sampling.
We use here the ResNet-18~\cite{he2016deep} backbone pretrained on ImageNet~\cite{ImageNet} and compare with TSM~\cite{lin2019tsm}.
We evalute slope clustering and cumulative clustering, and a cluster-free uniform grouping of evenly distributed segments and then aggregating them (\emph{Ours-uniform}).
\begin{table}[ht]
    \centering
    \resizebox{\linewidth}{!}{%
    \begin{tabular}{l r c c c c}
    \toprule
        \textbf{Model} &  \textbf{\#Frames} & \textbf{\#Clusters} & \textbf{FLOPs} & \textbf{Runtime} & \textbf{Top-1}\\ 
        &  && \textbf{/Video}& \textbf{Mem./Video} & \\ \midrule
        TSM \rule{0pt}{2.5ex}   & 8  & - & 14.56G &1.04GB &90.13 ± 0.38\\
        TSM \rule{0pt}{2.5ex}   & 16  & - & 29.12G & 1.72GB &  93.78 ± 0.33\\
        TSM  & all & - & 58.24G & 3.15GB &  \textbf{98.83 ± 0.16} \\\hline
        Ours-uniform \rule{0pt}{2.5ex} & all & 8 & 28.61G &1.56GB& 90.25 ± 0.28\\
        Ours-slope & all & 8 & 28.61G &1.56GB& 93.33 ± 0.19\\
        Ours-cumulative & all & 8 & 28.61G &1.56GB& \textbf{94.08 ± 0.25}\\\hline
        Ours-uniform \rule{0pt}{2.5ex} & all & 16 & 38.51G & 1.79GB & 92.73 ± 0.25\\
        Ours-slope & all & 16 & 38.51G & 1.79GB & 94.06 ± 0.18\\
        Ours-cumulative & all & 16 & 38.51G & 1.79GB & \textbf{95.27 ± 0.21}\\
    \bottomrule
    \end{tabular}}
    \vspace{0.05ex}
    \caption{Training with all frames gives best accuracy. 
        Our method with slope or cumulative clustering outperforms the uniform grouping of evenly distributed segments and frame sub-sampling. Our method has less FLOPs and runtime memory usage than TSM training with all frames.}
    \label{tab:Move4MNIST}
\end{table}

\tab{Move4MNIST} shows that TSM trained on all the 32 frames of a video in Move4MNIST significantly outperforms TSM trained on 8 and 16 sub-sampled frames. 
The uniform grouping of evenly distributed segments does not much better than random sub-sampling, and uniform grouping performs worse than random sub-sampling when the frame and cluster numbers increased from 8 to 16. This can be explained since the videos in the Move4MNIST contain black frames, UCF-101 background frames, and frames containing another digits at random positions, which can make sub-sampling miss frames related to the task and evenly distributed segments group frames wrongly. Both our clustering approaches with 8 and 16 clusters do better than evenly distributed segments or sub-sampling with 8 or 16 frames as they can adapt to the content and dynamically choose which frames to group.
In addition, our method has significantly reduced FLOPs and runtime memory when compared to the baseline on all frames. 

\begin{figure*}
    \centering
    \begin{tabular}{ccc}
    \small{Video 1} & \small{Video 2} & \small{Video 3} \\
    \includegraphics[width=0.325\linewidth]{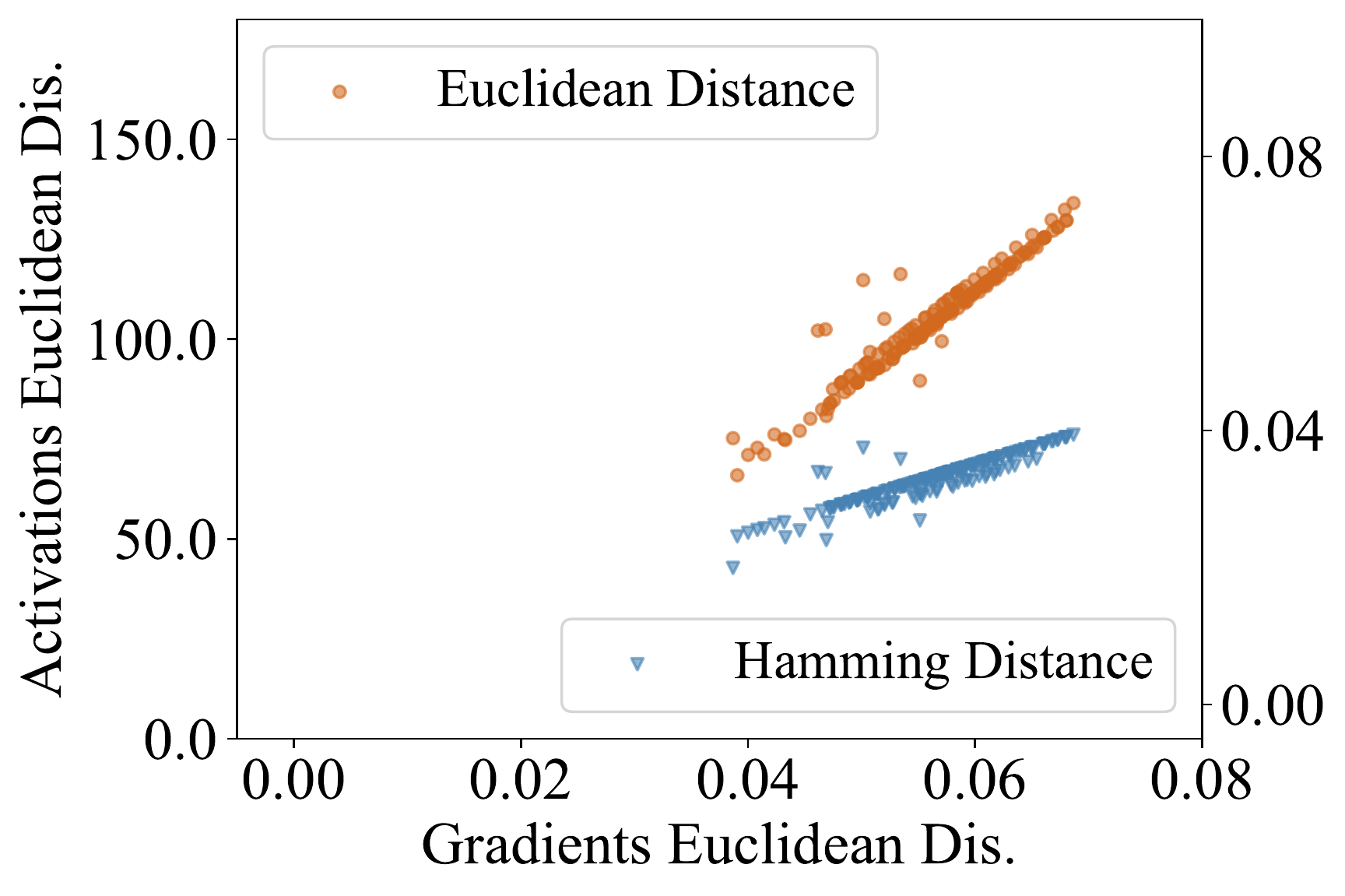} &  \includegraphics[width=0.312\linewidth]{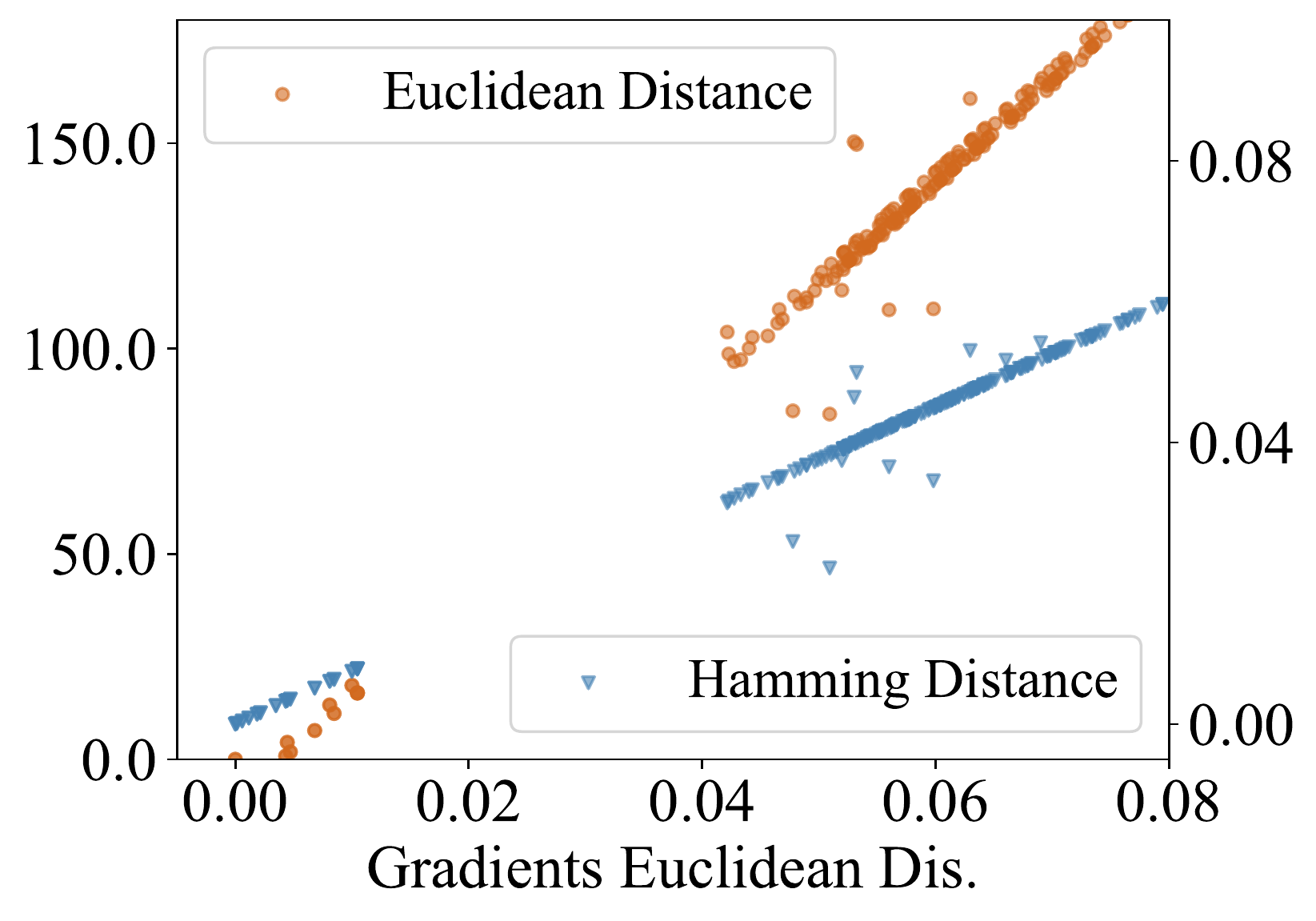} &
    \includegraphics[width=0.325\linewidth]{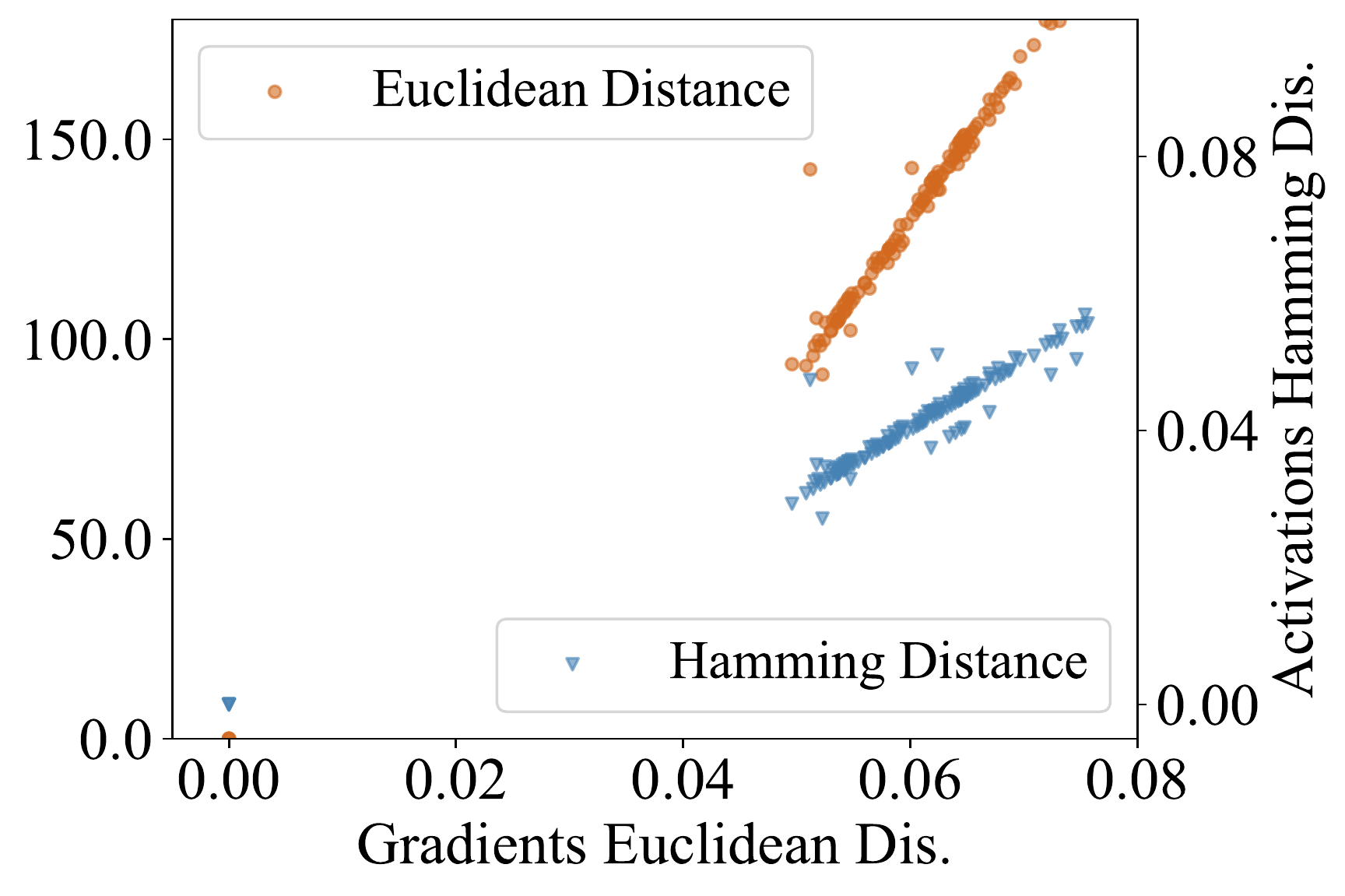}
    \end{tabular}
    \caption{An illustration of activation distance versus gradient distance for frames from three videos in Move4MNIST dataset. For frames that are similar with respect to recognizing the action, the activation distance and the gradients distance between them have a nearly linear relation for both the Euclidean distance and the Hamming distance.
    Our assumption that frames having similar activations with respect to the task have similar gradients is validated. }
    \label{fig:activ_grad}
\end{figure*}

%=====================================================================================
\subsection{Do similar frames have similar gradients?}

In this experiment, we evaluate our assumption that similar frame activations have similar gradients. The activations and gradients are taken from the 1st block of ResNet-18.
We show the Euclidean and the Hamming activation distance versus the gradient Euclidean distance between all $32*31/2=496$ frame pairs for three videos in Move4MNIST in \fig{activ_grad}.
For both the Euclidean distance and the Hamming distance the relation between activations and gradients is close to linear. It validates our assumption that frames having similar activations with respect to the task have similar gradients.

\begin{figure}
    \centering
      \includegraphics[width=\linewidth]{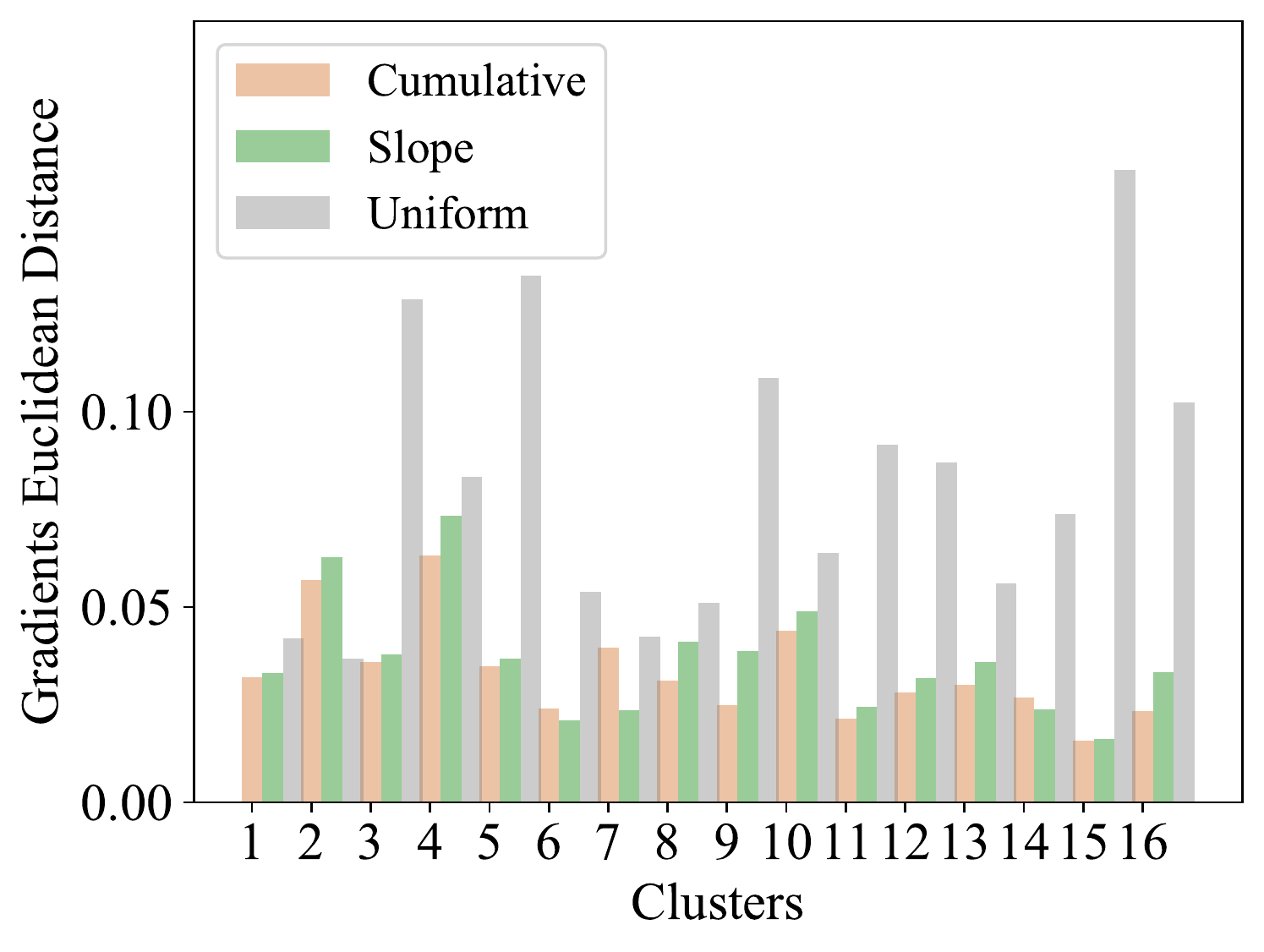}
    \caption{Comparing the Euclidean distance between gradients of the ground truth of truly using all frames to our efficient approximation per cluster for cumulative clustering, slope clustering and uniform grouping on Move4MNIST. Compared to uniform grouping and slope clustering, cumulative clustering results in smaller gradients difference and thus a better approximation.}
    \label{fig:grad_sum_grad}
\end{figure}

We compare the ground truth gradients when training truly on all frames to our efficient approximation.  We use 16 clusters and compare our approximate gradients to the real gradients which are from 3rd block of ResNet-18 for a video in Move4MNIST. We compare the results of our method with cumulative clustering, slope clustering and uniform grouping. Results in \fig{grad_sum_grad} show that compared to uniform grouping, cumulative clustering and slope clustering give smaller Euclidean distance between the single gradient from each cluster and the sum of gradients of frames in the corresponding cluster.
And cumulative clustering gives even smaller gradients differences than slope clustering.
In other words, it means that our method with cumulative clustering (the right hand side of \eq{ideal}) approximates the standard gradients calculation (the left hand side of \eq{ideal}) in the network with a small difference.

%=====================================================================================
\subsection{Analyzing model properties}
\begin{table}
    \centering
    \resizebox{\linewidth}{!}{%
    \begin{tabular}{l r r r l}
    \toprule
        \textbf{Model} &  \textbf{\#Frames} & \textbf{\#Clusters} & \textbf{Tr. sec/epoch} & \textbf{Top-1}\\ \hline
        TSM \rule{0pt}{2.5ex}   & 8  & - & 97.6 & 59.1\\
        TSM & 16  & - & 113.7 & 61.4\\ \hline
        Ours-uniform \rule{0pt}{2.5ex} & all & 8 & 100.1 & 58.3\\
        Ours-slope & all & 8 & 99.6 & 60.7\\
        Ours-cumulative & all & 8 & 101.3 & 62.0\\ \hline
        Ours-uniform \rule{0pt}{2.5ex} & all & 16 & 114.0 & 60.2\\
        Ours-slope & all & 16 & 114.5 & 63.7\\
        Ours-cumulative & all & 16 & 115.2 & \textbf{64.4}\\
    \bottomrule
    \end{tabular}
    }
    \vspace{0.05ex}
    \caption{With 8 and 16 clusters we consistently outperform TSM with 8 and 16 frames for comparable training time on the Breakfast dataset.}
    \label{tab:Breakfast}
\end{table}

\begin{figure}
    \centering
    \includegraphics[width=\linewidth]{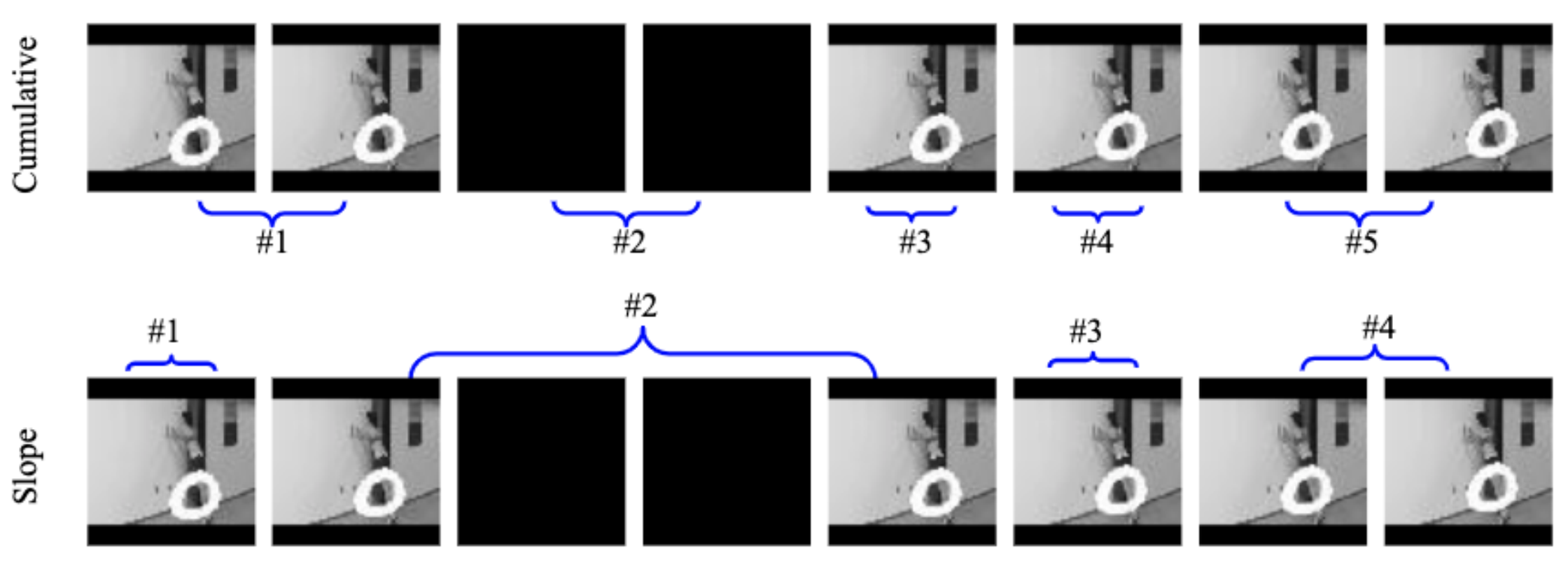}
    \caption{Temporal clustering results for a video in Move4MNIST. 
    Cumulative temporal clustering groups frames more accurately than slope temporal clustering.
    }
    \label{fig:clustering_compare}
\end{figure}

\begin{figure}
    \centering
    \includegraphics[width=\linewidth]{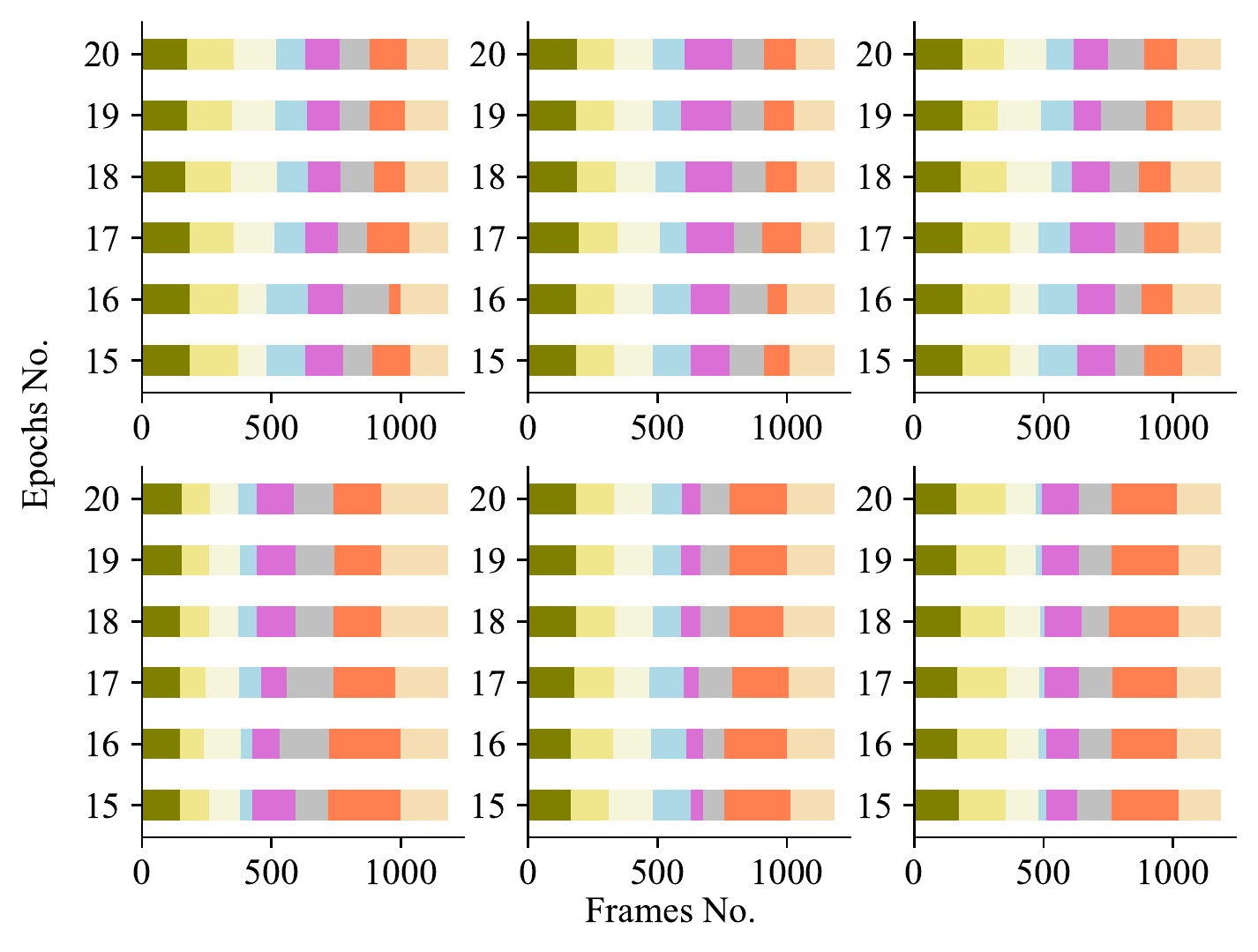}
    \caption{Cumulative temporal clustering results over epochs for six videos in the Breakfast dataset.
    Each cluster is shown in a different color. 
    Clusters contains segments with different lengths. 
    Our cumulative temporal clustering groups frames with similar activations together. 
    The cluster lengths change according to the changes in the frame activations during training.}
    \label{fig:clusters_visualization}
\end{figure}

We evaluate the clustering methods, the number of clusters, and the training time efficiency on Breakfast and Move4MNIST with a ResNet-18 backbone.

%---------------------------------------------------------------------------
\smallskip\noindent \textbf{Different temporal clustering methods.}
We compare slope clustering, cumulative clustering, and uniform grouping where the videos are split into equal segments. 
From~\tab{Breakfast}, cumulative clustering outperforms slope clustering, while uniform grouping has the lowest top-1 accuracy. This is because equal temporal grouping merges non similar frames together leading to linear approximations of non-linear information and incorrect network updates, resulting in a low action recognition accuracy. A similar trend is also visible on the Move4MNIST dataset in \tab{Move4MNIST}.

In \fig{clustering_compare}, we show the temporal clustering results for a small number of frames of a Move4MNIST video.
Cumulative clustering correctly groups similar frames together, while slope clustering groups moving zero frames and black frames together.

%---------------------------------------------------------------------------
\smallskip\noindent \textbf{Number of clusters.} 
We conduct experiments using 8 and 16 clusters for our method, which follows the protocol of TSM with 8 and 16 frames for training.
\tab{Breakfast} shows that using 16 clusters consistently outperforms using 8 clusters for all clustering methods.
A larger number of clusters improves accuracy. 
In the extreme case, the cluster numbers equal the number of frames in a video, which is equivalent with using all frames for training.
From the table we can also see that our cumulative temporal clustering implementation improves the top-1 accuracy by 2.9\% and 3.0\%, separately for 8 clusters and 16 clusters comparing to TSM with 8 and 16 frames.

To show that our cumulative temporal clustering algorithm is different from the naive uniform grouping, we visualize the 8 clusters obtained from cumulative temporal clustering for six videos over different epochs in the Breakfast dataset in \fig{clusters_visualization}.
Different videos have different segment lengths in the cumulative temporal clustering, which takes the similarity of frame activations into consideration.
In \fig{clusters_visualization}, we also show that the cluster length changes over epochs during training,
since the activations change during training.

%---------------------------------------------------------------------------
\smallskip\noindent \textbf{Efficiency of training time.}
\tab{Breakfast} gives the training time per epoch for all the models.
Our method with 8 clusters and 16 clusters only has an increase of 3.7 seconds and 1.5 seconds in training time per epoch, when compared to TSM with 8 frames and 16 frames.
The results show that our method is efficient during training time, while using all video frames.
%---------------------------------------------------------------------------
% \smallskip\noindent \textbf{Aggregation at different layers.}
% \begin{table}
%     \centering
%     \resizebox{\linewidth}{!}{%
%     \begin{tabular}{l c c c c}
%     \toprule
%         \textbf{Aggregation@} &  \textbf{FLOPs/Video} & \textbf{Runtime} & \textbf{Tr. sec} & \textbf{Top-1}\\
%         &  & \textbf{Mem./Video} & \textbf{/epoch} & \\ \midrule
%         Conv2\_x \rule{0pt}{2.5ex}   & 28.61G  & 1.56GB & 18.75& 94.08\%\\
%         Conv3\_x & 38.49G  & 1.95GB & 22.04 & 94.24\% \\ 
%         Conv4\_x  & 48.38G & 2.14GB & 25.94 & 94.41\%\\
%     \bottomrule
%     \end{tabular}
%     }
%     \caption{\small
%         Move4MNIST: Aggregation at different layers. 
%     }
%     %We perform our method with cumulative clustering on Move4MNIST using all 32 frames, and aggregated to 8 clusters at Conv2\_x, Conv3\_x and Conv4\_x block with ResNet18 backbone.
%     %Performing temporal aggregation at higher layers gives better accuracy but is slow and computationally and memory expensive.}
%     \label{tab:different layer}
% \end{table}
% \tab{different layer} shows that temporal aggregation at higher layers improves accuracy slightly at much increased computation and memory costs. Thus we aggregate at layer (Conv2\_x).

%=====================================================================================
\subsection{Comparison with the state-of-the-art}
We compare our method with the state-of-the-art on Something-Something V1\&V2, Breakfast, UCF-101 and HMDB51. All methods use ResNet-50 pre-trained on ImageNet as a backbone, unless specified otherwise.

%---------------------------------------------------------------------------
\smallskip\noindent \textbf{Comparison on the Breakfast dataset.}
\begin{table}
    \centering
    \resizebox{\linewidth}{!}{%
    \begin{tabular}{l l c c c}
    \toprule
        \textbf{Model} & \textbf{Backbone} & \textbf{\#Frames} & \textbf{\#Clusters} & \textbf{Top-1}\\ \midrule
        ResNet-152\cite{hussein2019videograph} \rule{0pt}{2.5ex} & ResNet152 & 64  & - &  41.1\%\\
        ActionVLAD~\cite{hussein2019videograph} & ResNet152 & 64 & - &  55.5\% \\
        VideoGraph~\cite{hussein2019videograph} & ResNet152 & 64 & - & 59.1\%\\
        TSM \cite{lin2019tsm} (our impl.) & ResNet50  & 16 & - & 72.1\%\\ \midrule
        Ours-slope & ResNet50 & all & 16 & 74.9\%\\
        Ours-cumulative & ResNet50 & all & 16 & \textbf{76.6\%}\\
    \bottomrule
    \end{tabular}}
    \vspace{0.05ex}
    \caption{Our method using either slope temporal clustering or cumulative temporal clustering compared to existing works on the Breakfast dataset. 
    Our proposal outperforms TSM, and significantly exceeds in top-1 accuracy methods using the deeper backbone architecture, ResNet-152.
    By using all frames our method has an advantage on long-term video action recognition.}
    \label{tab:SOTA-Breakfast}
\end{table}
We compare our method with existing work on the Breakfast dataset, which contains long action videos.
Our method using either slope temporal clustering or cumulative temporal clustering largely outperforms the three methods using ResNet-152 as a backbones, in \tab{SOTA-Breakfast}.
Compared to TSM using 16 sub-sampled frames, our method improves the top-1 accuracy by 2.8\% and 4.5\% with slope temporal clustering and cumulative temporal clustering, respectively.
Methods using sub-sampling can easily miss important frames for the recognition task on long action videos.
Our method has an advantage on the long videos for action recognition, by efficiently utilizing all the frames.

%---------------------------------------------------------------------------
\begin{table*}
    \centering
    \resizebox{\linewidth}{!}{%
    \begin{tabular}{l l l c c c c}
    \toprule
        \textbf{Model} & \textbf{Backbone} & \textbf{Pre-train} &\textbf{\#Frames} & \textbf{\#Clusters} & \textbf{Top-1 UCF-101} & \textbf{Top-1 HMDB51}\\ \hline
        TSM~\cite{lin2019tsm} (our impl.) \rule{0pt}{2.5ex} & ResNet50 & Kinetics & 1 & - & 91.2\% & 65.1\%\\
        TSN~\cite{lin2019tsm}    & ResNet50  & Kinetics &  8 & - & 91.7\%& 64.7\%\\
        SI+DI+OF+DOF~\cite{BilenDynamicImage2018}  & ResNeXt50 & Imagenet &  dynamic images & - & 95.0\% & 71.5\% \\
        TSM~\cite{lin2019tsm} & ResNet50 & Kinetics & 8 & - & 95.9\% & 73.5\%\\ 
        STM~\cite{STM} & ResNet50 & ImageNet+Kinetics & 16 & - & 96.2\% & 72.2\%\\ \hline
        Ours-{slope} \rule{0pt}{2.5ex} & TSM-ResNet50 & Kinetics & all & 8 & 96.2\% & 73.3\%\\
        Ours-{cumulative} & TSM-ResNet50 & Kinetics & all & 8 & \textbf{96.4\%} &\textbf{73.4\%}\\
    \bottomrule
    \end{tabular}}
    \vspace{0.05ex}
    \caption{Top-1 accuracy on UCF-101 and HMDB51.
    Our method performs only slightly better than the state-of-the-art on the scene-related datasets UCF-101 and HMDB51. 
    These datasets do not have much frame diversity per video, thus, the improvement of our method over sampling methods is limited.
    }
    \label{tab:UCF101andHMDB51}
\end{table*}

\smallskip\noindent \textbf{Comparison on the Something-Something dataset.}
\begin{table}
    \centering
    \resizebox{\linewidth}{!}{%
    \begin{tabular}{l r c c c}
    \toprule
        \textbf{Model} &  \textbf{\#Frames} & \textbf{\#Clusters} & \textbf{Top-1 V1} & \textbf{Top-1 V2}\\ \hline
        TSN~\cite{lin2019tsm} \rule{0pt}{2.5ex}   & 8  & - & 19.7\%& 30.0\%\\
        TRN-Multiscale~\cite{lin2019tsm} & 8  & - & 38.9\% & 48.8\% \\ 
        TSM~\cite{lin2019tsm}  & 8 & - & 45.6\% & 59.1\%\\
        TSM~\cite{STM}  & 16 & - & 47.2\% & 63.4\%\\
        STM~\cite{STM} & 8 & - & 49.2\% & 62.3\%\\
        STM~\cite{STM} & 16 & - & 50.7\% & 64.2\%\\ \hline
        Ours-slope \rule{0pt}{2.5ex} & all & 8 & 46.7\% & 60.2\%\\
        Ours-cumulative & all & 8 & 49.5\% & 62.7\%\\
        Ours-cumulative & all & 16 & \textbf{51.4}\% & \textbf{65.1\%}\\
    \bottomrule
    \end{tabular}
    }
    \vspace{0.05ex}
    \caption{Top-1 accuracy on Something-Something V1 and V2 datasets. 
    Our method using cumulative temporal clustering outperforms the state-of-the-art methods on both Something-Something V1 and V2. 
    Our method achieves limited accuracy improvement for shorter videos.}
    \label{tab:somethin-Something}
\end{table}
In \tab{somethin-Something}, we list the results of our method compared to other methods on the Something-Something V1\&V2 datasets.
We achieve state-of-the-art performance on both V1 and V2, with outperforming STM of 8 frames by 0.3\% and 0.4\% for V1 and V2, and STM of 16 frames by 0.7\% and 0.9\% for V1 and V2 respectively.
Comparing to TSM, we significantly improve the top-1 accuracy of 8 frames by 3.9\% and 3.5\%, and the top-1 accuracy of 16 frames by 4.2\% and 1.7\% for the V1 and V2 datasets.
Although the Something-Something dataset is characterized by temporal variations, the video clips are short compared to the Breakfast dataset.
The methods using frame sampling heuristics can capture the main movement in videos.
Therefore, our accuracy improvement is not as pronounced as for the Breakfast dataset.

%---------------------------------------------------------------------------
\smallskip\noindent \textbf{Comparison on the UCF-101 and HMDB51 datasets.} We train with 8 clusters and evaluate over three splits and report averaged results in~\tab{UCF101andHMDB51}.
Our performance is on par with state-of-the-art methods on both datasets.
The UCF-101 and HMDB51 have a scene-bias, where motion plays a limited role and just a few number of frames --or even a single frame-- is sufficient. Thus, methods relying on sampling heuristics can correctly classify the actions and our method using all frames is not expected to improve results.  
To test this, we show results with a single frame in \tab{UCF101andHMDB51}  which shows that TSM with 1 frame achieves comparable accuracy to TSN with 8 frames on UCF-101 and outperforms TSN with 8 frames on HMDB51. 
For scene-biased datasets, using all frames does not bring accuracy benefits.

\section{Conclusion}
We propose an efficient method for training action recognition deep networks without relying on sampling heuristics.
Our work offers a solution to using all video frames during training based on the assumption that similar frames have similar gradients, leading to similar parameter updates.
To this end, we efficiently find frames that are similar with respect to the classification task, by using a cumulative temporal clustering algorithm based on Hamming distances.
The clustering based on Hamming distances enforces that activations in a cluster agree in signs, which is a requirement entailed by our assumption that we can approximate the gradients of multiple frames with a single gradient of an aggregated frame.
We accumulate the activations within each cluster to create new representations used to classify the actions.
Our proposed method shows competitive results on large datasets when compared to existing work. 

Despite our state of the art results, we identify several limitations. One limitation is that the number of clusters is fixed and thus not well-suited for inhomogeneous videos with more semantic (shot) changes than clusters. This could create a dependency for action proposals or other approaches to pre-segment a video in homogeneous segments which somewhat counters the philosophy of using full video action recognition.  Another limitation is that for grouping frames the only non-linearity we consider is the activation function and do not use the non-linearity in the loss. This limitation seems insurmountable, as memory constraints prevent us to store all frame activations for when the loss is computed. Nevertheless, with our current results and analysis, we make a first move for action recognition to go full video.\\

{\small \noindent \textbf{Acknowledgments}
This work is part of the research program Efficient Deep Learning (EDL), which is (partly) financed by the Dutch Research Council (NWO). }

{\small
    \bibliographystyle{ieee_fullname}
    % \bibliography{ieee}

}

% \clearpage
\onecolumn
\section*{Appendix}
In addition to the comparison with 2D models, we also show results of our method compared to the state-of-the-art 3D models and additional 2D models on Breakfast, Something-Something V1 \& V2, UCF-101 and HMDB51.
[Nx] denotes the new citations in the tables. 
% Our model is more computationally efficient during than 3D models.
\subsection*{Comparison on the Breakfast dataset.}
\begin{table*}[h]
    \centering
    \resizebox{\textwidth}{!}{%
    \begin{tabular}{l l c c c c c}
    \toprule
        \textbf{Model} & \textbf{Backbone} & \textbf{\#3D} & \textbf{\#Optical flow} & \textbf{\#Frames} & \textbf{\#Clusters} & \textbf{Top-1}\\ \midrule
        ResNet152\cite{hussein2019videograph} \rule{0pt}{2.5ex} & ResNet152 &- & - & 64  & - &  41.1\%\\
        ActionVLAD~\cite{hussein2019videograph} & ResNet152 & -& - & 64 & - &  55.5\% \\
        VideoGraph~\cite{hussein2019videograph} & ResNet152 & - & -& 64 & - & 59.1\%\\
        TSM \cite{lin2019tsm} (our impl.) & ResNet50 & - & - & 16 & - & 72.1\%\\ \midrule
        I3D~\cite{hussein2019videograph} & 3D Inception-v1 & \checkmark & - & 512 & - & 58.6\% \\
        I3D + ActionVLAD~\cite{hussein2019videograph} & 3D Inception-v1 & \checkmark & - & 512 & - & 65.5\% \\
         I3D + VideoGraph~\cite{hussein2019videograph} & 3D Inception-v1 & \checkmark & - & 512 & - & 69.5\% \\
        3D ResNet-50 + Timeception~\cite{hussein2019timeception} & 3D ResNet-50 & \checkmark & - & 512 & - & 71.3\% \\ \midrule
        Ours-slope & ResNet50 & - & - & all & 16 & 74.9\%\\
        Ours-cumulative & ResNet50 & - & - & all & 16 & \textbf{76.6\%}\\
    \bottomrule
    \end{tabular}}
    \vspace{0.05ex}
    \caption{Our method using either slope temporal clustering or cumulative temporal clustering compared to existing works on the Breakfast dataset. 
    Our proposal outperforms TSM and the 3D model, and significantly exceeds in top-1 accuracy methods using the deeper backbone architecture, ResNet-152.
    By using all frames our method has an advantage on long-term video action recognition.}
    \label{tab:SOTA-Breakfast-sub}
\end{table*}
%====================================================================================================
\subsection*{Comparison on the Something-Something dataset.}
\begin{table*}[h]
    \centering
    \resizebox{\textwidth}{!}{%
    \begin{tabular}{l l c c c c c c}
    \toprule
        \textbf{Model} &  \textbf{Backbone} & \textbf{\#3D} & \textbf{\#Optical flow} & \textbf{\#Frames} & \textbf{\#Clusters} & \textbf{Top-1 V1} & \textbf{Top-1 V2}\\ \hline
        TSN~\cite{lin2019tsm} \rule{0pt}{2.5ex} & ResNet50 & - & - & 8  & - & 19.7\%& 30.0\%\\
        TRN-Multiscale~\cite{lin2019tsm} & ResNet50 & - & - & 8  & - & 38.9\% & 48.8\% \\ 
        TSM~\cite{lin2019tsm} & Resnet50 & - & - & 8 & - & 45.6\% & 59.1\%\\
        STM~\cite{STM} & ResNet50 & - & - & 8 & - & 49.2\% & 62.3\%\\ 
        MSNet-R50~\citeN{Kwon2020Motionsqueeze} & TSM-ResNet50 & - & - & 8 & - & \textbf{50.9\%} & \textbf{63.0\%}\\ \hline
        I3D~\citeN{Wang2018SpaceTime} \rule{0pt}{2.5ex} & I3D & \checkmark & - & 32 & - & 41.6\% & - \\
        NL-I3D~\citeN{Wang2018SpaceTime} & I3D & \checkmark & - & 32 & - & 44.4\% & -\\
        NL-I3D+GCN~\citeN{Wang2018SpaceTime} & I3D & \checkmark & - & 32 & - & 46.1\% & -\\
        S3D-G ~\citeN{Xie2018S3D-G} & Inception & \checkmark & - & 64 & - & 48.2\% & -\\
        ECO~\cite{zolfaghari2018eco} & BNIncep+3D Res18 & \checkmark & - & 8 & -& 39.6\% & - \\
        ECO~\cite{zolfaghari2018eco} & BNIncep+3D Res18 & \checkmark & - & 16 & - & 41.4\% & - \\
        ECO-{En} \textit{Lite}~\cite{zolfaghari2018eco} & BNIncep+3D Res18 & \checkmark & - & 92 & - & 46.4\% & - \\
        ECO-{En} \textit{Lite}-{RGB+Flow}~\cite{zolfaghari2018eco} & BNIncep$+$3D Res18 & \checkmark & \checkmark & 92+92 & - & 49.5\% & - \\ 
        DFB-Net~\citeN{Martinez2019DFB-Net} & 3D ResNet50 & \checkmark & - & 16 & - & 50.1\% & -\\ \hline
        Ours-{slope} \rule{0pt}{2.5ex} & TSM-ResNet50 & - & - & all & 8 & 46.7\% & 60.2\%\\
        Ours-{cumulative} & TSM-ResNet50 & - & - & all & 8 & 49.5\% & 62.7\%\\
    \bottomrule
    \end{tabular}}
    \vspace{0.05ex}
    \caption{Top-1 accuracy on Something-Something V1 and V2 datasets. 
    Our method using cumulative temporal clustering outperforms most state-of-the-art methods on both Something-Something V1 and V2, and performs on par with ECO-{En} \textit{Lite} with both RGB and optical flow while slightly worse than MSNet-R50.
    Our method achieves limited accuracy improvement for shorter videos.}
    \label{tab:somethin-Something-sub}
\end{table*}
%======================================================================================================
\newpage
\subsection*{Comparison on the UCF-101 and HMDB51 dataset.}
\begin{table*}[h]
    \centering
    \resizebox{\textwidth}{!}{%
    \begin{tabular}{l l l c c c c c c}
    \toprule
        \textbf{Model} & \textbf{Backbone} & \textbf{Pre-train} & \textbf{\#3D} & \textbf{\#Optical flow} &\textbf{\#Frames} & \textbf{\#Clusters} & \textbf{Top-1 UCF-101} & \textbf{Top-1 HMDB51}\\ \hline
        TSM~\cite{lin2019tsm} (our impl.) \rule{0pt}{2.5ex} & ResNet50 & Kinetics & - & - & 1 & - & 91.2\% & 65.1\%\\
        TSN~\cite{lin2019tsm}    & ResNet50  & Kinetics & - & - &  8 & - & 91.7\%& 64.7\%\\
        SI+DI+OF+DOF~\cite{BilenDynamicImage2018}  & ResNeXt50 & Imagenet & - & \checkmark &  dynamic images & - & 95.0\% & 71.5\% \\
        TSM~\cite{lin2019tsm} & ResNet50 & Kinetics & - & - & 8 & - & 95.9\% & 73.5\%\\ 
        STM~\cite{STM} & ResNet50 & ImageNet+Kinetics & - & - & 16 & - & 96.2\% & 72.2\%\\
        MSNet-R50~\citeN{Kwon2020Motionsqueeze} & TSM-ResNet50 & Kinetics & - & - & 8 & - & - & 75.8\%\\ \hline
        ECO-{En} \textit{Lite}~\cite{zolfaghari2018eco} \rule{0pt}{2.5ex} & BNIncep+3D Res18 & Kinetics & \checkmark & - & 8  &- & 94.8\% &72.4\%\\
        RGB I3D~\cite{carreira2017quo} & 3D Inception-v1 & Kinetics & \checkmark & - & 64 & - & 95.1\% & 74.3\%\\
        Two-stream I3D~\cite{carreira2017quo} & 3D Inception-v1 & Kinetics & \checkmark & \checkmark & 64+64 & - & \textbf{97.8\%} & \textbf{80.9\%}\\ \hline 
        Ours-{slope} \rule{0pt}{2.5ex} & TSM-ResNet50 & Kinetics & - & - & all & 8 & 96.2\% & 73.3\% \\ 
        Ours-{cumulative} & TSM-ResNet50 & Kinetics & - & - & all & 8 &  96.4\% & 73.4\%\\
    \bottomrule
 \end{tabular}
 }
    \vspace{0.05ex}
    \caption{ Top-1 accuracy on UCF-101 and HMDB51.
    Our method performs only slightly better than the state-of-the-art on the scene-related datasets UCF-101 and HMDB51, and worse than two-stream I3D, which uses both RGB and optical flow with 3D Inception-v1 backbone.
    Given that these datasets do not have a large number of frames per video, the improvement of our method over sampling methods is limited.}
    \label{tab:UCF101andHMDB51-sub}
\end{table*}

{\small
    \bibliographyN{ieee_sub}
    % \bibliographystyleN{ieee_fullname}

}

\end{document}